# AN OPTIMAL TABULAR PARSING ALGORITHM

**Mark-Jan Nederhof** [*]
University of Nijmegen, Department of Computer Science
Toernooiveld, 6525 ED Nijmegen, The Netherlands
markjan@cs.kun.nl

## Abstract

In this paper we relate a number of parsing algorithms which have been developed in very different areas of parsing theory, and which include deterministic algorithms, tabular algorithms, and a parallel algorithm. We show that these algorithms are based on the same underlying ideas.

By relating existing ideas, we hope to provide an opportunity to improve some algorithms based on features of others. A second purpose of this paper is to answer a question which has come up in the area of tabular parsing, namely how to obtain a parsing algorithm with the property that the table will contain as little entries as possible, but without the possibility that two entries represent the same subderivation.

## Introduction

Left-corner (LC) parsing is a parsing strategy which has been used in different guises in various areas of computer science. Deterministic LC parsing with $k$ symbols of lookahead can handle the class of LC($k$) grammars. Since LC parsing is a very simple parsing technique and at the same time is able to deal with left recursion, it is often used as an alternative to top-down (TD) parsing, which cannot handle left recursion and is generally less efficient.

Nondeterministic LC parsing is the foundation of a very efficient parsing algorithm [7], related to Tomita's algorithm and Earley's algorithm. It has one disadvantage however, which becomes noticeable when the grammar contains many rules whose right-hand sides begin with the same few grammars symbols, e.g.

$$A \rightarrow \alpha\beta_1 \mid \alpha\beta_2 \mid \ldots$$

where $\alpha$ is not the empty string. After an LC parser has recognized the first symbol $X$ of such an $\alpha$, it will as next step predict all aforementioned rules. This amounts to much nondeterminism, which is detrimental both to the time-complexity and the space-complexity.

[*]Supported by the Dutch Organisation for Scientific Research (NWO), under grant 00-62-518

A first attempt to solve this problem is to use predictive LR (PLR) parsing. PLR parsing allows simultaneous processing of a common prefix $\alpha$, provided that the left-hand sides of the rules are the same. However, in case we have e.g. the rules $A \rightarrow \alpha\beta_1$ and $B \rightarrow \alpha\beta_2$, where again $\alpha$ is not the empty string but now $A \neq B$, then PLR parsing will not improve the efficiency. We therefore go one step further and discuss extended LR (ELR) and common-prefix (CP) parsing, which are algorithms capable of simultaneous processing of all common prefixes. ELR and CP parsing are the foundation of tabular parsing algorithms and a parallel parsing algorithm from the existing literature, but they have not been described in their own right.

To the best of the author's knowledge, the various parsing algorithms mentioned above have not been discussed together in the existing literature. The main purpose of this paper is to make explicit the connections between these algorithms.

A second purpose of this paper is to show that CP and ELR parsing are obvious solutions to a problem of tabular parsing which can be described as follows. For each parsing algorithm working on a stack there is a realisation using a parse table, where the parse table allows sharing of computation between different search paths. For example, Tomita's algorithm [18] can be seen as a tabular realisation of nondeterministic LR parsing.

At this point we use the term *state* to indicate the symbols occurring on the stack of the original algorithm, which also occur as entries in the parse table of its tabular realisation.

In general, powerful algorithms working on a stack lead to efficient tabular parsing algorithms, provided the grammar can be handled almost deterministically. In case the stack algorithm is very nondeterministic for a certain grammar however, sophistication which increases the number of states may lead to an increasing number of entries in the parse table of the tabular realization. This can be informally explained by the fact that each state represents the computation of a number of subderivations. If the number of states is increased then it is inevitable that at some point some states represent an overlapping collection of subderivations, which may lead to work being repeated during parsing. Furthermore, the parse forest (a compact representation of all parse trees) which is output by a tabular algorithm may in this case not be optimally dense.

We conclude that we have a tradeoff between the case that the grammar allows almost deterministic parsing and the case that the stack algorithm is very nondeterministic for a certain grammar. In the former case, sophistication leads to *less* entries in the table, and in the latter case, sophistication leads to *more* entries, provided this sophistication is realised by an increase in the number of states. This is corroborated by empirical data from [1, 4], which deal with tabular LR parsing.

As we will explain, CP and ELR parsing are more deterministic than most other parsing algorithms for many grammars, but their tabular realizations can never compute the same subderivation twice. This represents an optimum in a range of possible parsing algorithms.

This paper is organized as follows. First we discuss nondeterministic left-corner parsing, and demonstrate how common prefixes in a grammar may be a source of bad performance for this technique.

Then, a multitude of parsing techniques which exhibit better treatment of common prefixes is discussed. These techniques, including nondeterministic PLR, ELR, and CP parsing, have their origins in theory of deterministic, parallel, and tabular parsing. Subsequently, the application to parallel and tabular parsing is investigated more closely.

Further, we briefly describe how rules with empty right-hand sides complicate the parsing process.

The ideas described in this paper can be generalized to head-driven parsing, as argued in [9].

We will take some liberty in describing algorithms from the existing literature, since using the original descriptions would blur the similarities of the algorithms to one another. In particular, we will not treat the use of lookahead, and we will consider all algorithms working on a stack to be nondeterministic. We will only describe *recognition* algorithms. Each of the algorithms can however be easily extended to yield parse trees as a side-effect of recognition.

The notation used in the sequel is for the most part standard and is summarised below.

A context-free grammar $G = (T, N, P, S)$ consists of two finite disjoint sets $N$ and $T$ of nonterminals and terminals, respectively, a start symbol $S \in N$, and a finite set of rules $P$. Every rule has the form $A \to \alpha$, where the left-hand side (lhs) $A$ is an element from $N$ and the right-hand side (rhs) $\alpha$ is an element from $V^*$, where $V$ denotes $(N \cup T)$. $P$ can also be seen as a relation on $N \times V^*$.

We use symbols $A, B, C, \ldots$ to range over $N$, symbols $a, b, c, \ldots$ to range over $T$, symbols $X, Y, Z$ to range over $V$, symbols $\alpha, \beta, \gamma, \ldots$ to range over $V^*$, and $v, w, x, \ldots$ to range over $T^*$. We let $\epsilon$ denote the empty string. The notation of rules $A \to \alpha_1, A \to \alpha_2, \ldots$ with the same lhs is often simplified to $A \to \alpha_1 | \alpha_2 | \ldots$

A rule of the form $A \to \epsilon$ is called an *epsilon rule*. We assume grammars do not have epsilon rules unless stated otherwise.

The relation $P$ is extended to a relation $\to$ on $V^* \times V^*$ as usual. The reflexive and transitive closure of $\to$ is denoted by $\to^*$.

We define: $B \angle A$ if and only if $A \to B\alpha$ for some $\alpha$. The reflexive and transitive closure of $\angle$ is denoted by $\angle^*$, and is called the *left-corner relation*.

We say two rules $A \to \alpha_1$ and $B \to \alpha_2$ have a *common prefix* $\beta$ if $\alpha_1 = \beta\gamma_1$ and $\alpha_2 = \beta\gamma_2$, for some $\gamma_1$ and $\gamma_2$, where $\beta \neq \epsilon$.

A recognition algorithm can be specified by means of a push-down automaton $\mathcal{A} = (T, Alph, Init, \vdash, Fin)$, which manipulates configurations of the form $(\Gamma, v)$, where $\Gamma \in Alph^*$ is the stack, constructed from left to right, and $v \in T^*$ is the remaining input.

The initial configuration is $(Init, w)$, where $Init \in Alph$ is a distinguished stack symbol, and $w$ is the input. The steps of an automaton are specified by means of the relation $\vdash$. Thus, $(\Gamma, v) \vdash (\Gamma', v')$ denotes that $(\Gamma', v')$ is obtainable from $(\Gamma, v)$ by one step of the automaton. The reflexive and transitive closure of $\vdash$ is denoted by $\vdash^*$. The input $w$ is accepted if $(Init, w) \vdash^* (Fin, \epsilon)$, where $Fin \in Alph$ is a distinguished stack symbol.

## LC parsing

For the definition of left-corner (LC) recognition [7] we need stack symbols (*items*) of the form $[A \to \alpha \bullet \beta]$, where $A \to \alpha\beta$ is a rule, and $\alpha \neq \epsilon$. (Remember that we do not allow epsilon rules.) The informal meaning of an item is "The part before the dot has just been recognized, the first symbol after the dot is to be recognized next". For technical reasons we also need the items $[S' \to \bullet S]$ and $[S' \to S \bullet]$, where $S'$ is a fresh symbol. Formally:

$$I^{LC} = \{[A \to \alpha \bullet \beta] \mid A \to \alpha\beta \in P^\dagger \wedge (\alpha \neq \epsilon \vee A = S')\}$$

where $P^\dagger$ represents the *augmented* set of rules, consisting of the rules in $P$ plus the extra rule $S' \to S$.

**Algorithm 1 (Left-corner)**
$\mathcal{A}^{LC} = (T, I^{LC}, Init, \vdash, Fin)$, $Init = [S' \to \bullet S]$, $Fin = [S' \to S \bullet]$. Transitions are allowed according to the following clauses.

1. $(\Gamma[B \to \beta \bullet C\gamma], av) \vdash$
   $\qquad (\Gamma[B \to \beta \bullet C\gamma][A \to a \bullet \alpha], v)$
   where there is $A \to a\alpha \in P^\dagger$ such that $A \angle^* C$
2. $(\Gamma[A \to \alpha \bullet a\beta], av) \vdash (\Gamma[A \to \alpha a \bullet \beta], v)$
3. $(\Gamma[B \to \beta \bullet C\gamma][A \to \alpha \bullet], v) \vdash$
   $\qquad (\Gamma[B \to \beta \bullet C\gamma][D \to A \bullet \delta], v)$
   where there is $D \to A\delta \in P^\dagger$ such that $D \angle^* C$
4. $(\Gamma[B \to \beta \bullet A\gamma][A \to \alpha \bullet], v) \vdash (\Gamma[B \to \beta A \bullet \gamma], v)$

The conditions using the left-corner relation $\angle^*$ in the first and third clauses together form a feature which is called *top-down (TD) filtering*. TD filtering makes sure that subderivations that are being computed bottom-up may eventually grow into subderivations with the required root. TD filtering is not necessary for a correct algorithm, but it reduces nondeterminism, and guarantees the *correct-prefix property*, which means that in case of incorrect input the parser does not read past the first incorrect character.

**Example 1** Consider the grammar with the following rules:

$$\begin{aligned} E &\rightarrow E+T \mid T \uparrow E \mid T \\ T &\rightarrow T*F \mid T**F \mid F \\ F &\rightarrow a \end{aligned}$$

It is easy to see that $E \angle E, T \angle E, T \angle T, F \angle T$. The relation $\angle^*$ contains $\angle$ but from the reflexive closure it also contains $F \angle^* F$ and from the transitive closure it also contains $F \angle^* E$.

The recognition of $a*a$ is realised by:

|   |                                                          | $a*a$ |
|---|----------------------------------------------------------|-------|
| 1 | $[E' \rightarrow \bullet E]$                             |       |
| 1 | $[E' \rightarrow \bullet E][F \rightarrow a \bullet]$    | $*a$  |
| 2 | $[E' \rightarrow \bullet E][T \rightarrow F \bullet]$    | $*a$  |
| 3 | $[E' \rightarrow \bullet E][T \rightarrow T \bullet *F]$ | $*a$  |
| 4 | $[E' \rightarrow \bullet E][T \rightarrow T * \bullet F]$| $a$   |
| 5 | $[E' \rightarrow \bullet E][T \rightarrow T * \bullet F][F \rightarrow a \bullet]$ | |
| 6 | $[E' \rightarrow \bullet E][T \rightarrow T * F \bullet]$| |
| 7 | $[E' \rightarrow \bullet E][E \rightarrow T \bullet]$    | |
| 8 | $[E' \rightarrow E \bullet]$                             | |

Note that since the automaton does not use any lookahead, Step 3 may also have replaced $[T \rightarrow F \bullet]$ by any other item besides $[T \rightarrow T \bullet *F]$ whose rhs starts with $T$ and whose lhs satisfies the condition of top-down filtering with regard to $E$, i.e. by $[T \rightarrow T \bullet **F]$, $[E \rightarrow T \bullet \uparrow E]$, or $[E \rightarrow T \bullet]$. □

LC parsing with $k$ symbols of lookahead can handle deterministically the so called LC($k$) grammars. This class of grammars is formalized in [13].[1] How LC parsing can be improved to handle common *suffixes* efficiently is discussed in [6]; in this paper we restrict our attention to common *prefixes*.

## PLR, ELR, and CP parsing

In this section we investigate a number of algorithms which exhibit a better treatment of common prefixes.

### Predictive LR parsing

Predictive LR (PLR) parsing with $k$ symbols of lookahead was introduced in [17] as an algorithm which yields efficient parsers for a subset of the LR($k$) grammars [16] and a superset of the LC($k$) grammars. How deterministic PLR parsing succeeds in handling a larger class of grammars (the PLR($k$) grammars) than the LC($k$) grammars can be explained by identifying PLR parsing for some grammar $G$ with LC parsing for some grammar $G'$ which results after applying a transformation called *left-factoring*.

---
[1] In [17] a different definition of the LC($k$) grammars may be found, which is not completely equivalent.

Left-factoring consists of replacing two or more rules $A \rightarrow \alpha\beta_1 \mid \alpha\beta_2 \mid \ldots$ with a common prefix $\alpha$ by the rules $A \rightarrow \alpha A'$ and $A' \rightarrow \beta_1 \mid \beta_2 \mid \ldots$, where $A'$ is a fresh nonterminal. The effect on LC parsing is that a choice between rules is postponed until after all symbols of $\alpha$ are completely recognized. Investigation of the next $k$ symbols of the remaining input may then allow a choice between the rules to be made deterministically.

The PLR algorithm is formalised in [17] by transforming a PLR($k$) grammar into an LL($k$) grammar and then assuming the standard realisation of LL($k$) parsing. When we consider nondeterministic top-down parsing instead of LL($k$) parsing, then we obtain the new formulation of nondeterministic PLR(0) parsing below.

We first need to define another kind of item, viz. of the form $[A \rightarrow \alpha]$ such that there is at least one rule of the form $A \rightarrow \alpha\beta$ for some $\beta$. Formally:

$$I^{PLR} = \{[A \rightarrow \alpha] \mid A \rightarrow \alpha\beta \in P^\dagger \wedge (\alpha \neq \epsilon \vee A = S')\}$$

Informally, an item $[A \rightarrow \alpha] \in I^{PLR}$ represents one or more items $[A \rightarrow \alpha \bullet \beta] \in I^{LC}$.

**Algorithm 2 (Predictive LR)**
$A^{PLR} = (T, I^{PLR}, Init, \vdash, Fin)$, $Init = [S' \rightarrow ]$, $Fin = [S' \rightarrow S]$, and $\vdash$ defined by:

1. $(\Gamma[B \rightarrow \beta], av) \vdash (\Gamma[B \rightarrow \beta][A \rightarrow a], v)$
   where there are $A \rightarrow a\alpha, B \rightarrow \beta C\gamma \in P^\dagger$ such that $A \angle^* C$

2. $(\Gamma[A \rightarrow \alpha], av) \vdash (\Gamma[A \rightarrow \alpha a], v)$
   where there is $A \rightarrow \alpha a\beta \in P^\dagger$

3. $(\Gamma[B \rightarrow \beta][A \rightarrow \alpha], v) \vdash (\Gamma[B \rightarrow \beta][D \rightarrow A], v)$
   where $A \rightarrow \alpha \in P^\dagger$ and where there are $D \rightarrow A\delta, B \rightarrow \beta C\gamma \in P^\dagger$ such that $D \angle^* C$

4. $(\Gamma[B \rightarrow \beta][A \rightarrow \alpha], v) \vdash (\Gamma[B \rightarrow \beta A], v)$
   where $A \rightarrow \alpha \in P^\dagger$ and where there is $B \rightarrow \beta A\gamma \in P^\dagger$

**Example 2** Consider the grammar from Example 1. Using Predictive LR, recognition of $a*a$ is realised by:

|   |                                              | $a*a$ |
|---|----------------------------------------------|-------|
|   | $[E' \rightarrow ]$                          |       |
| 1 | $[E' \rightarrow ][F \rightarrow a]$         | $*a$  |
| 2 | $[E' \rightarrow ][T \rightarrow F]$         | $*a$  |
| 3 | $[E' \rightarrow ][T \rightarrow T]$         | $*a$  |
| 4 | $[E' \rightarrow ][T \rightarrow T*]$        | $a$   |
| ⋮ | ⋮                                            | ⋮     |
| 8 | $[E' \rightarrow E]$                         |       |

Comparing these configurations with those reached by the LC recognizer, we see that here after Step 3 the stack element $[T \rightarrow T]$ represents both $[T \rightarrow T \bullet *F]$ and $[T \rightarrow T \bullet **F]$, so that nondeterminism is reduced. Still some nondeterminism remains, since Step 3 could also have replaced $[T \rightarrow F]$ by $[E \rightarrow T]$, which represents both $[E \rightarrow T \bullet \uparrow E]$ and $[E \rightarrow T \bullet]$. □

## Extended LR parsing

An *extended* context-free grammar has right-hand sides consisting of arbitrary regular expressions over $V$. This requires an LR parser for an extended grammar (an *ELR* parser) to behave differently from normal LR parsers.

The behaviour of a normal LR parser upon a reduction with some rule $A \rightarrow \alpha$ is very simple: it pops $|\alpha|$ states from the stack, revealing, say, state $Q$; it then pushes state $goto(Q, A)$. (We identify a state with its corresponding set of items.)

For extended grammars the behaviour upon a reduction cannot be realised in this way since the regular expression of which the rhs is composed may describe strings of various lengths, so that it is unknown how many states need to be popped.

In [11] this problem is solved by forcing the parser to decide at each call $goto(Q, X)$ whether

**a)** $X$ is one more symbol of an item in $Q$ of which some symbols have already been recognized, or whether

**b)** $X$ is the first symbol of an item which has been introduced in $Q$ by means of the closure function.

In the second case, a state which is a variant of $goto(Q, X)$ is *pushed* on top of state $Q$ as usual. In the first case, however, state $Q$ on top of the stack is *replaced* by a variant of $goto(Q, X)$. This is safe since we will never need to return to $Q$ if after some more steps we succeed in recognizing some rule corresponding with one of the items in $Q$. A consequence of the action in the first case is that upon reduction we need to pop only one state off the stack.

Further work in this area is reported in [5], which treats nondeterministic ELR parsing and therefore does not regard it as an obstacle if a choice between cases **a)** and **b)** cannot be uniquely made.

We are not concerned with extended context-free grammars in this paper. However, a very interesting algorithm results from ELR parsing if we restrict its application to ordinary context-free grammars. (We will maintain the name "*extended* LR" to stress the origin of the algorithm.) This results in the new nondeterministic ELR(0) algorithm that we describe below, derived from the formulation of ELR parsing in [5].

First, we define a set of items as

$$I = \{[A \rightarrow \alpha \bullet \beta] \mid A \rightarrow \alpha\beta \in P^\dagger\}$$

Note that $I^{LC} \subset I$. If we define for each $Q \subseteq I$:

$closure(Q) =$
$\quad Q \cup \{[A \rightarrow \bullet \alpha] \mid [B \rightarrow \beta \bullet C\gamma] \in Q \wedge A \angle^* C\}$

then the $goto$ function for LR(0) parsing is defined by

$goto(Q, X) =$
$\quad closure(\{[A \rightarrow \alpha X \bullet \beta] \mid [A \rightarrow \alpha \bullet X\beta] \in Q\})$

For ELR parsing however, we need two $goto$ functions, $goto_1$ and $goto_2$, one for kernel items (i.e. those in $I^{LC}$) and one for nonkernel items (the others). These are defined by

$goto_1(Q, X) =$
$\quad closure(\{[A \rightarrow \alpha X \bullet \beta] \mid [A \rightarrow \alpha \bullet X\beta] \in Q \wedge$
$\quad\quad\quad\quad\quad (\alpha \neq \epsilon \vee A = S')\})$

$goto_2(Q, X) =$
$\quad closure(\{[A \rightarrow X \bullet \beta] \mid [A \rightarrow \bullet X\beta] \in Q \wedge A \neq S'\})$

At each shift (where $X$ is some terminal) and each reduce with some rule $A \rightarrow \alpha$ (where $X$ is $A$) we may nondeterministically apply $goto_1$, which corresponds with case **a)**, or $goto_2$, which corresponds with case **b)**. Of course, one or both may not be defined on $Q$ and $X$, because $goto_i(Q, X)$ may be $\emptyset$, for $i \in \{1, 2\}$.

Now remark that when using $goto_1$ and $goto_2$, each reachable set of items contains only items of the form $A \rightarrow \alpha \bullet \beta$, for some fixed string $\alpha$, plus some nonkernel items. We will ignore the nonkernel items since they can be derived from the kernel items by means of the closure function.

This suggests representing each set of items by a new kind of item of the form $[\{A_1, A_2, \ldots, A_n\} \rightarrow \alpha]$, which represents all items $A \rightarrow \alpha \bullet \beta$ for some $\beta$ and $A \in \{A_1, A_2, \ldots, A_n\}$. Formally:

$I^{ELR} = \{[\Delta \rightarrow \alpha] \mid \emptyset \subset \Delta \subseteq \{A \mid A \rightarrow \alpha\beta \in P^\dagger\} \wedge$
$\quad\quad\quad\quad (\alpha \neq \epsilon \vee \Delta = \{S'\})\}$

where we use the symbol $\Delta$ to range over sets of nonterminals.

**Algorithm 3 (Extended LR)**
$A^{ELR} = (T, I^{ELR}, Init, \vdash, Fin)$, $Init = [\{S'\} \rightarrow ]$, $Fin = [\{S'\} \rightarrow S]$, and $\vdash$ defined by:

1. $(\Gamma[\Delta \rightarrow \beta], av) \vdash (\Gamma[\Delta \rightarrow \beta][\Delta' \rightarrow a], v)$
   where $\Delta' = \{A \mid \exists A \rightarrow a\alpha, B \rightarrow \beta C\gamma \in P^\dagger [B \in \Delta \wedge A \angle^* C]\}$ is non-empty

2. $(\Gamma[\Delta \rightarrow \alpha], av) \vdash (\Gamma[\Delta' \rightarrow \alpha a], v)$
   where $\Delta' = \{A \in \Delta \mid A \rightarrow \alpha a\beta \in P^\dagger\}$ is non-empty

3. $(\Gamma[\Delta \rightarrow \beta][\Delta' \rightarrow \alpha], v) \vdash (\Gamma[\Delta \rightarrow \beta][\Delta'' \rightarrow A], v)$
   where there is $A \rightarrow \alpha \in P^\dagger$ with $A \in \Delta'$, and $\Delta'' = \{D \mid \exists D \rightarrow A\delta, B \rightarrow \beta C\gamma \in P^\dagger[B \in \Delta \wedge D \angle^* C]\}$ is non-empty

4. $(\Gamma[\Delta \rightarrow \beta][\Delta' \rightarrow \alpha], v) \vdash (\Gamma[\Delta'' \rightarrow \beta A], v)$
   where there is $A \rightarrow \alpha \in P^\dagger$ with $A \in \Delta'$, and $\Delta'' = \{B \in \Delta \mid B \rightarrow \beta A\gamma \in P^\dagger\}$ is non-empty

Note that Clauses 1 and 3 correspond with $goto_2$ and that Clauses 2 and 4 correspond with $goto_1$.

**Example 3** Consider again the grammar from Example 1. Using the ELR algorithm, recognition of $a * a$ is realised by:

|   |   | $a * a$ |
|---|---|---|
|   | $[\{E'\} \rightarrow ]$ | $a * a$ |
| 1 | $[\{E'\} \rightarrow ][\{F\} \rightarrow a]$ | $* a$ |
| 2 | $[\{E'\} \rightarrow ][\{T\} \rightarrow F]$ | $* a$ |
| 3 | $[\{E'\} \rightarrow ][\{T, E\} \rightarrow T]$ | $* a$ |
| 4 | $[\{E'\} \rightarrow ][\{T\} \rightarrow T *]$ | $a$ |
| $\vdots$ | $\vdots$ | $\vdots$ |
| 8 | $[\{E'\} \rightarrow E]$ | |

Comparing these configurations with those reached by the PLR recognizer, we see that here after Step 3 the stack element $[\{T,E\} \to T]$ represents both $[T \to T \bullet * F]$ and $[T \to T \bullet * * F]$, but also $[E \to T \bullet]$ and $[E \to T \bullet \uparrow E]$, so that nondeterminism is even further reduced. □

A simplified ELR algorithm, which we call the *pseudo* ELR algorithm, results from avoiding reference to $\Delta$ in Clauses 1 and 3. In Clause 1 we then have a simplified definition of $\Delta'$, viz. $\Delta' = \{A \mid \exists A \to a\alpha, B \to \beta C\gamma \in P^\dagger [A \angle^* C]\}$, and in the same way we have in Clause 3 the new definition $\Delta'' = \{D \mid \exists D \to A\delta, B \to \beta C\gamma \in P^\dagger [D \angle^* C]\}$. Pseudo ELR parsing can be more easily realised than full ELR parsing, but the correct-prefix property can no longer be guaranteed. Pseudo ELR parsing is the foundation of a tabular algorithm in [20].

## Common-prefix parsing

One of the more complicated aspects of the ELR algorithm is the treatment of the sets of nonterminals in the left-hand sides of items. A drastically simplified algorithm is the basis of a tabular algorithm in [21]. Since in [21] the algorithm itself is not described but only its tabular realisation,[2] we take the liberty of giving this algorithm our own name: *common-prefix (CP) parsing*, since it treats all rules with a common prefix simultaneously.[3]

The simplification consists of omitting the sets of nonterminals in the left-hand sides of items:

$$I^{CP} = \{[\to \alpha] \mid A \to \alpha\beta \in P^\dagger\}$$

**Algorithm 4 (Common-prefix)**
$A^{CP} = (T, I^{CP}, Init, \vdash, Fin)$, $Init = [\to]$, $Fin = [\to S]$, and $\vdash$ defined by:

1. $(\Gamma[\to \beta], av) \vdash (\Gamma[\to \beta][\to a], v)$
   where there are $A \to a\alpha, B \to \beta C\gamma \in P^\dagger$ such that $A \angle^* C$

2. $(\Gamma[\to \alpha], av) \vdash (\Gamma[\to \alpha a], v)$
   where there is $A \to \alpha a\beta \in P^\dagger$

3. $(\Gamma[\to \beta][\to \alpha], v) \vdash (\Gamma[\to \beta][\to A], v)$
   where there are $A \to \alpha, D \to A\delta, B \to \beta C\gamma \in P^\dagger$ such that $D \angle^* C$

4. $(\Gamma[\to \beta][\to \alpha], v) \vdash (\Gamma[\to \beta A], v)$
   where there are $A \to \alpha, B \to \beta A\gamma \in P^\dagger$

The simplification which leads to the CP algorithm inevitably causes the correct-prefix property to be lost.

**Example 4** Consider again the grammar from Example 1. It is clear that $a + a \uparrow a$ is not a correct string according to this grammar. The CP algorithm may go through the following sequence of configurations:

|    |                                                    |                         |
|----|----------------------------------------------------|-------------------------|
| 1  | $[\to][\to a]$                                     | $a + a \uparrow a$      |
| 2  | $[\to][\to F]$                                     | $+ a \uparrow a$        |
| 3  | $[\to][\to T]$                                     | $+ a \uparrow a$        |
| 4  | $[\to][\to E]$                                     | $+ a \uparrow a$        |
| 5  | $[\to][\to E +]$                                   | $a \uparrow a$          |
| 6  | $[\to][\to E +][\to a]$                            | $\uparrow a$            |
| 7  | $[\to][\to E +][\to F]$                            | $\uparrow a$            |
| 8  | $[\to][\to E +][\to T]$                            | $\uparrow a$            |
| 9  | $[\to][\to E +][\to T \uparrow]$                   | $a$                     |
| 10 | $[\to][\to E +][\to T \uparrow][\to a]$            |                         |

| | $[\to]$ | $a + a \uparrow a$ |

We see that in Step 9 the first incorrect symbol $\uparrow$ is read, but recognition then continues. Eventually, the recognition process is blocked in some unsuccessful configuration, which is guaranteed to happen for any incorrect input[4]. In general however, after reading the first incorrect symbol, the algorithm may perform an unbounded number of steps before it halts. (Imagine what happens for input of the form $a + a \uparrow a + a + a + \ldots + a$.) □

## Tabular parsing

Nondeterministic push-down automata can be realised efficiently using parse tables [1]. A parse table consists of sets $T_{i,j}$ of items, for $0 \leq i \leq j \leq n$, where $a_1 \ldots a_n$ represents the input. The idea is that an item is only stored in a set $T_{i,j}$ if the item represents recognition of the part of the input $a_{i+1} \ldots a_j$.

We will first discuss a tabular form of CP parsing, since this is the most simple parsing technique discussed above. We will then move on to the more difficult ELR technique. Tabular PLR parsing is fairly straightforward and will not be discussed in this paper.

### Tabular CP parsing

CP parsing has the following tabular realization:

**Algorithm 5 (Tabular common-prefix)**
Sets $T_{i,j}$ of the table are to be subsets of $I^{CP}$. Start with an empty table. Add $[\to]$ to $T_{0,0}$. Perform one of the following steps until no more items can be added.

1. Add $[\to a]$ to $T_{i-1,i}$ for $a = a_i$ and $[\to \beta] \in T_{j,i-1}$ where there are $A \to a\alpha, B \to \beta C\gamma \in P^\dagger$ such that $A \angle^* C$

2. Add $[\to \alpha a]$ to $T_{j,i}$ for $a = a_i$ and $[\to \alpha] \in T_{j,i-1}$ where there is $A \to \alpha a\beta \in P^\dagger$

3. Add $[\to A]$ to $T_{j,i}$ for $[\to \alpha] \in T_{j,i}$ and $[\to \beta] \in T_{h,j}$ where there are $A \to \alpha, D \to A\delta, B \to \beta C\gamma \in P^\dagger$ such that $D \angle^* C$

4. Add $[\to \beta A]$ to $T_{h,i}$ for $[\to \alpha] \in T_{j,i}$ and $[\to \beta] \in T_{h,j}$ where there are $A \to \alpha, B \to \beta A\gamma \in P^\dagger$

Report recognition of the input if $[\to S] \in T_{0,n}$.

For an example, see Figure 1.

Tabular CP parsing is related to a variant of CYK parsing with TD filtering in [5]. A form of tabular

---

[2]An attempt has been made in [19] but this paper does not describe the algorithm in its full generality.

[3]The original algorithm in [21] applies an optimization concerning unit rules, irrelevant to our discussion.

[4]unless the grammar is cyclic, in which case the parser may not terminate, both on correct and on incorrect input

|   | 0 | 1 | 2 | 3 | 4 | 5 |
|---|---|---|---|---|---|---|
| 0 | $[\to]$ (0) | $[\to a]$ (1) $[\to F]$ (2) $[\to T]$ (3) $[\to E]$ (4) | $[\to E+]$(5) | $[\to E+T]$ $[\to E]$ | $\emptyset$ | $\emptyset$ |
| 1 |   |   | $\emptyset$ | $\emptyset$ | $\emptyset$ | $\emptyset$ |
| 2 |   |   |   | $[\to a]$ (6) $[\to F]$ (7) $[\to T]$ (8) | $[\to T\uparrow]$ (9) | $[\to T\uparrow E]$ |
| 3 |   |   |   |   | $\emptyset$ | $\emptyset$ |
| 4 |   |   |   |   |   | $[\to a]$ (10) $[\to F]$ $[\to T]$ $[\to E]$ |

Figure 1: Tabular CP parsing

Consider again the grammar from Example 1 and the (incorrect) input $a + a \uparrow a$. After execution of the tabular common-prefix algorithm, the table is as given here. The sets $T_{j,i}$ are given at the $j$-th row and $i$-th column.
The items which correspond with those from Example 4 are labelled with $(0), (1), \ldots$ These labels also indicate the order in which these items are added to the table.

CP parsing without top-down filtering (i.e. without the checks concerning the left-corner relation $\angle^*$) is the main algorithm in [21].

Without the use of top-down filtering, the references to $[\to \beta]$ in Clauses 1 and 3 are clearly not of much use any more. When we also remove the use of these items, then these clauses become:

1. Add $[\to a]$ to $T_{i-1,i}$      for $a = a_i$
   where there is $A \to a\alpha \in P^\dagger$
3. Add $[\to A]$ to $T_{j,i}$      for $[\to \alpha] \in T_{j,i}$
   where there are $A \to \alpha, D \to A\delta \in P^\dagger$

In the resulting algorithm, no set $T_{i,j}$ depends on any set $T_{g,h}$ with $g < i$. In [15] this fact is used to construct a parallel parser with $n$ processors $P_0, \ldots, P_{n-1}$, with each $P_i$ processing the sets $T_{j,i}$ for all $j > i$. The flow of data is strictly from right to left, i.e. items computed by $P_i$ are only passed on to $P_0, \ldots, P_{i-1}$.

### Tabular ELR parsing

The tabular form of ELR parsing allows an optimization which constitutes an interesting example of how a tabular algorithm can have a property not shared by its nondeterministic origin.[5]

First note that we can compute the columns of a parse table strictly from left to right, that is, for fixed $i$ we can compute all sets $T_{j,i}$ before we compute the sets $T_{j,i+1}$.

If we formulate a tabular ELR algorithm in a naive way analogously to Algorithm 5, as is done in [5], then for example the first clause is given by:

1. Add $[\Delta' \to a]$ to $T_{i-1,i}$ for $a = a_i$ and $[\Delta \to \beta] \in T_{j,i-1}$
   where $\Delta' = \{A \mid \exists A \to a\alpha, B \to \beta C\gamma \in P^\dagger [B \in \Delta \land A \angle^* C]\}$ is non-empty

---
[5]This is reminiscent of the *admissibility tests* [3], which are applicable to *tabular* realisations of logical push-down automata, but not to these automata themselves.

However, for certain $i$ there may be many $[\Delta \to \beta] \in T_{j,i-1}$, for some $j$, and each may give rise to a different $\Delta'$ which is non-empty. In this way, Clause 1 may add several items $[\Delta' \to a]$ to $T_{i-1,i}$, some possibly with overlapping sets $\Delta'$. Since items represent computation of subderivations, the algorithm may therefore compute the same subderivation several times.

We propose an optimization which makes use of the fact that all possible items $[\Delta \to \beta] \in T_{j,i-1}$ are already present when we compute items in $T_{i-1,i}$: we compute one single item $[\Delta' \to a]$, where $\Delta'$ is a large set computed using all $[\Delta \to \beta] \in T_{j,i-1}$, for any $j$. A similar optimization can be made for the third clause.

**Algorithm 6 (Tabular extended LR)**
Sets $T_{i,j}$ of the table are to be subsets of $I^{ELR}$. Start with an empty table. Add $[\{S'\} \to ]$ to $T_{0,0}$. For $i = 1, \ldots, n$, in this order, perform one of the following steps until no more items can be added.

1. Add $[\Delta' \to a]$ to $T_{i-1,i}$      for $a = a_i$
   where $\Delta' = \{A \mid \exists j \exists [\Delta \to \beta] \in T_{j,i-1} \exists A \to a\alpha, B \to \beta C\gamma \in P^\dagger [B \in \Delta \land A \angle^* C]\}$ is non-empty
2. Add $[\Delta' \to \alpha a]$ to $T_{j,i}$      for $a = a_i$ and
   $[\Delta \to \alpha] \in T_{j,i-1}$
   where $\Delta' = \{A \in \Delta \mid A \to \alpha a\beta \in P^\dagger\}$ is non-empty
3. Add $[\Delta'' \to A]$ to $T_{j,i}$      for $[\Delta' \to \alpha] \in T_{j,i}$
   where there is $A \to \alpha \in P^\dagger$ with $A \in \Delta'$, and $\Delta'' = \{D \mid \exists h \exists [\Delta \to \beta] \in T_{h,j} \exists D \to A\delta, B \to \beta C\gamma \in P^\dagger [B \in \Delta \land D \angle^* C]\}$ is non-empty
4. Add $[\Delta'' \to \beta A]$ to $T_{h,i}$      for $[\Delta' \to \alpha] \in T_{j,i}$ and
   $[\Delta \to \beta] \in T_{h,j}$
   where there is $A \to \alpha \in P^\dagger$ with $A \in \Delta'$, and $\Delta'' = \{B \in \Delta \mid B \to \beta A\gamma \in P^\dagger\}$ is non-empty

Report recognition of the input if $[\{S'\} \to S] \in T_{0,n}$.

Informally, the top-down filtering in the first and third clauses is realised by investigating all left corners $D$ of nonterminals $C$ (i.e. $D \angle^* C$) which are expected

from a certain input position. For input position $i$ these nonterminals $D$ are given by

$$S_i = \{D \mid \exists j \exists [\Delta \to \beta] \in T_{j,i} \\ \exists B \to \beta C \gamma \in P^{\dagger}[B \in \Delta \wedge D \angle^* C]\}$$

Provided each set $S_i$ is computed just after completion of the $i$-th column of the table, the first and third clauses can be simplified to:

1. Add $[\Delta' \to a]$ to $T_{i-1,i}$     for $a = a_i$
where $\Delta' = \{A \mid A \to a\alpha \in P^{\dagger}\} \cap S_{i-1}$ is non-empty

3. Add $[\Delta'' \to A]$ to $T_{j,i}$     for $[\Delta' \to \alpha] \in T_{j,i}$
where there is $A \to \alpha \in P^{\dagger}$ with $A \in \Delta'$, and $\Delta'' = \{D \mid D \to A\delta \in P^{\dagger}\} \cap S_j$ is non-empty which may lead to more practical implementations.

Note that we may have that the tabular ELR algorithm manipulates items of the form $[\Delta \to \alpha]$ which would not occur in any search path of the nondeterministic ELR algorithm, because in general such a $\Delta$ is the union of many sets $\Delta'$ of items $[\Delta' \to \alpha]$ which would be manipulated at the same input position by the nondeterministic algorithm in *different* search paths.

With minor differences, the above tabular ELR algorithm is described in [21]. A tabular version of *pseudo* ELR parsing is presented in [20]. Some useful data structures for practical implementation of tabular and non-tabular PLR, ELR and CP parsing are described in [8].

### Finding an optimal tabular algorithm

In [14] Schabes derives the LC algorithm from LR parsing similar to the way that ELR parsing can be derived from LR parsing. The LC algorithm is obtained by not only splitting up the goto function into $goto_1$ and $goto_2$ but also splitting up $goto_2$ even further, so that it nondeterministically yields the closure of one single kernel item. (This idea was described earlier in [5], and more recently in [10].)

Schabes then argues that the LC algorithm can be determinized (i.e. made more deterministic) by manipulating the goto functions. One application of this idea is to take a fixed grammar and choose different goto functions for different parts of the grammar, in order to tune the parser to the grammar.

In this section we discuss a different application of this idea: we consider various goto functions which are *global*, i.e. which are the same for all parts of a grammar. One example is ELR parsing, as its $goto_2$ function can be seen as a determinized version of the $goto_2$ function of LC parsing. In a similar way we obtain PLR parsing. Traditional LR parsing is obtained by taking the full determinization, i.e. by taking the normal goto function which is not split up.[6]

We conclude that we have a family consisting of LC, PLR, ELR, and LR parsing, which are increasingly deterministic. In general, the more deterministic an algorithm is, the more parser states it requires. For example, the LC algorithm requires a number of states (the items in $I^{LC}$) which is linear in the size of the grammar. By contrast, the LR algorithm requires a number of states (the sets of items) which is *exponential* in the size of the grammar [2].

The differences in the number of states complicates the choice of a tabular algorithm as the one giving optimal behaviour for all grammars. If a grammar is very simple, then a sophisticated algorithm such as LR may allow completely deterministic parsing, which requires a linear number of entries to be added to the parse table, measured in the size of the grammar.

If, on the other hand, the grammar is very ambiguous such that even LR parsing is very nondeterministic, then the tabular realisation may at worst add each state to each set $T_{i,j}$, so that the more states there are, the more work the parser needs to do. This favours simple algorithms such as LC over more sophisticated ones such as LR. Furthermore, if more than one state represents the same subderivation, then computation of that subderivation may be done more than once, which leads to parse forests (compact representations of collections of parse trees) which are not optimally dense [1, 12, 7].

Schabes proposes to tune a parser to a grammar, or in other words, to use a *combination* of parsing techniques in order to find an optimal parser for a certain grammar.[7] This idea has until now not been realised. However, when we try to find a single parsing algorithm which performs well for all grammars, then the tabular ELR algorithm we have presented may be a serious candidate, for the following reasons:

- For all $i$, $j$, and $\alpha$ at most one item of the form $[\Delta \to \alpha]$ is added to $T_{i,j}$. Therefore, identical subderivations are not computed more than once. (This is a consequence of our optimization in Algorithm 6.) Note that this also holds for the tabular CP algorithm.

- ELR parsing guarantees the correct-prefix property, contrary to the CP algorithm. This prevents computation of all subderivations which are useless with regard to the already processed input.

- ELR parsing is more deterministic than LC and PLR parsing, because it allows shared processing of all common prefixes. It is hard to imagine a practical parsing technique more deterministic than ELR parsing which also satisfies the previous two properties. In particular, we argue in [8] that refinement of the LR technique in such a way that the first property above holds would require an impractically large number of LR states.

---

[6]Schabes more or less also argues that LC itself can be obtained by determinizing TD parsing. (In lieu of TD parsing he mentions Earley's algorithm, which is its tabular realisation.)

[7]This is reminiscent of the idea of "optimal cover" [5].

## Epsilon rules

Epsilon rules cause two problems for bottom-up parsing. The first is non-termination for simple realisations of nondeterminism (such as backtrack parsing) caused by *hidden left recursion* [7]. The second problem occurs when we optimize TD filtering e.g. using the sets $S_i$: it is no longer possible to completely construct a set $S_i$ before it is used, because the computation of a derivation deriving the empty string requires $S_i$ for TD filtering but at the same time its result causes new elements to be added to $S_i$. Both problems can be overcome [8].

## Conclusions

We have discussed a range of different parsing algorithms, which have their roots in compiler construction, expression parsing, and natural language processing. We have shown that these algorithms can be described in a common framework.

We further discussed tabular realisations of these algorithms, and concluded that we have found an optimal algorithm, which in most cases leads to parse tables containing fewer entries than for other algorithms, but which avoids computing identical subderivations more than once.

## Acknowledgements

The author acknowledges valuable correspondence with Klaas Sikkel, René Leermakers, François Barthélemy, Giorgio Satta, Yves Schabes, and Frédéric Voisin.